\documentclass[10pt, conference, letterpaper]{IEEEtran}

\IEEEoverridecommandlockouts
\usepackage{cite}
\usepackage{amsmath,amssymb,amsfonts}
\usepackage{algorithm}
\usepackage{algorithmic}
\usepackage{graphicx}
\usepackage{textcomp}
\usepackage{subfigure}
\usepackage{xcolor}
\def\BibTeX{{\rm B\kern-.05em{\sc i\kern-.025em b}\kern-.08em
    T\kern-.1667em\lower.7ex\hbox{E}\kern-.125emX}}
\begin{document}

\title{BSDGAN: Balancing Sensor Data Generative Adversarial Networks for Human Activity Recognition\\
}

\author{\IEEEauthorblockN{1\textsuperscript{st} Yifan Hu}
\IEEEauthorblockA{\textit{Department of Computer Science and Technology} \\
\textit{Nanjing Tech University}\\
Nanjing, China \\
Email: liquid0707@outlook.com}
\and
\IEEEauthorblockN{2\textsuperscript{nd} Yu Wang}
\IEEEauthorblockA{\textit{School of Computer Science and Technology} \\
\textit{Xidian University}\\
Xi’an, China \\
Email: wangyu@njtech.edu.cn}
}

\maketitle

\begin{abstract}
The development of IoT technology enables a variety of sensors can be integrated into mobile devices. Human Activity Recognition (HAR) based on sensor data has become an active research topic in the field of machine learning and ubiquitous computing. However, due to the inconsistent frequency of human activities, the amount of data for each activity in the human activity dataset is imbalanced.
Considering the limited sensor resources and the high cost of manually labeled sensor data, human activity recognition is facing the challenge of highly imbalanced activity datasets.

In this paper, we propose Balancing Sensor Data Generative Adversarial Networks (BSDGAN) to generate sensor data for minority human activities. The proposed BSDGAN consists of a generator model and a discriminator model. Considering the extreme imbalance of human activity dataset, an autoencoder is employed to initialize the training process of BSDGAN, ensure the data features of each activity can be learned. The generated activity data is combined with the original dataset to balance the amount of activity data across human activity classes.
We deployed multiple human activity recognition models on two publicly available imbalanced human activity datasets, WISDM and UNIMIB. Experimental results show that the proposed BSDGAN can effectively capture the data features of real human activity sensor data, and generate realistic synthetic sensor data. Meanwhile, the balanced activity dataset can effectively help the activity recognition model to improve the recognition accuracy.
\end{abstract}

\begin{IEEEkeywords}
Human activity recognition, generative adversarial networks, data augmentation
\end{IEEEkeywords}

\section{Introduction}
With the popularization of smart mobile devices, the research on human activity recognition based on sensor data occupies an increasing proportion in the field of ubiquitous computing\cite{jobanputra2019human}. The miniaturization of sensors enables mobile devices to embed various sensors and collect various types of sensor data, including acceleration, gyroscope, magnetic field strength and pressure. The abundance of sensor data leads to Human Activity Recognition (HAR) based on sensor data widely employed in various fields of our daily life, such as health care\cite{wang2019survey}, sleep quality analysis\cite{sathyanarayana2016robust}, patient rehabilitation\cite{schrader2020advanced}, interactive entertainment\cite{bulling2014tutorial} and sports fitness\cite{lockhart2012applications}. The successful deployment of human activity recognition shows great development potential and research significance.

The collection and label of sensor data is the basis of the entire human activity recognition research. The classification accuracy of a human activity recognition model depends on the dataset itself\cite{wu2016mixed}. The performance of human activity recognition classifier is affected by an extreme imbalance in the amount of sensor data between activity classes\cite{chen2017performance}. However, most HAR researchers only focus on the accuracy of human activity recognition rather than the data quality of the human activity dataset.
The vast majority of sensor data in human activity datasets come from mobile smartphones and wearable devices\cite{chen2021deep}. Considering the purchase cost of smart devices and the labor cost of collecting and labeling sensor data, most human activity datasets have a highly imbalanced data amount between activity classes, the research of human activity recognition is facing the challenge of data amount imbalance.

In recent years, many data generation methods based on deep learning have been proposed. Among them, Generative Adversarial Networks (GAN) is the most effective and most interesting data generation method. The original GANs framework was firstly proposed by Ian Goodfellow et al. \cite{goodfellow2016nips} in 2014 and consisted of a generator and discriminator built by multilayer perceptron (MLP).
GAN learns the data feature distribution of real data through the adversarial between the generator and the discriminator, enabling the generator model can generate fake data to fool the discriminator.
In the field of computer vision, GAN has been proved that it can effectively generate many types of high-quality images, including human faces, animals and landscapes\cite{bao2017cvae}. Nevertheless, few researchers have successfully deployed independent GANs on human activity datasets to generate high-quality sensor data because of the huge data feature difference between different activity classes\cite{hamad2020joint}.

In this paper, we employed a unified independent generative adversarial network, BSDGAN, to improve the performance of human activity recognition models by oversampling human activity classes with small amounts of sensor data.
The main contributions of this paper are summarized as follows:\\
\indent(1) We successfully deploy the proposed generative adversarial networks on imbalanced human activity datasets, and generate sensor data for specified human activities.\\
\indent(2) We added conditional constraints to the GAN framework and improved the loss function with gradient penalty. An autoencoder is employed to initialize the GAN training, which gives the BSDGAN data feature distribution knowledge of all human activities, helps stabilize the training process.\\
\indent(3) We oversample the human activity classes with few data to balance the dataset, prove that the balanced human activity dataset can improve the performance of HAR models.

The rest of this paper is organized as follows. Section II reviews the related works regarding HAR and Generative Adversarial Networks. Section III presents the details of our proposed BSDGAN. Section IV illustrates the performance of BSDGAN on imbalanced human activity datasets and presents the comparison of various HAR models before and after data balancing. Section V gives the conclusion of this paper and introduces our future work.

\section{Related Work}

\subsection{Human Activity Recognition}

Human activity recognition is divided into visual image-based activity recognition\cite{caba2015activitynet} and sensor data-based activity recognition\cite{de2018sensor}. Compared with human activity recognition methods based on visual images, sensor-based methods have the advantages of small data amount, simple reasoning, low cost and strong embeddability, occupying a major position in the research and application of human activity recognition. With the popularity of smartphones, human activity recognition based on smartphones has been applied in many research fields. Some researchers employed traditional machine learning methods to recognize human activities, such as K-Nearest Neighbors\cite{paul2015effective}, Decision Tree\cite{fan2013human} and Random Forest\cite{feng2015random}. In recent years, with the rapid development of deep learning, many researchers employed deep learning methods to perform human activity recognition.

Traditional machine learning methods always require handcrafted data features. In order to achieve desired recognition results, researchers have devoted a lot of effort to researching and designing effective features to improve HAR performance\cite{ali2018depth,yala2015feature}. In contrast, deep learning methods can automatically extract features from sensor data\cite{wang2019deep}. Song-Mi Lee et al.\cite{lee2017human} used a one-dimensional Convolutional Neural Network (CNN) to analyze the three-axis data of the accelerometer. They convert the three-axis data into vector magnitude data to achieve high classification accuracy. C.A.Ronaoo et al.\cite{Ronaoo2015EvaluationOD} used a CNN with two hidden layers to analyze accelerometer data and gyroscope data, achieved activity classification accuracy of over 90\% on the University of California's Open Activity Dataset (UCI DATASET)\cite{anguita2013public}. H. Zhang et al.\cite{zhang2019novel} used a multi-head CNN combined with attention mechanism to recognize various activities, including walking, standing, sitting, jogging, going upstairs and downstairs. Y. Chen et al.\cite{chen2016lstm} used a Long Short-Term Memory neural network (LSTM) to analyze the sensor data. R. Mutegeki et al.\cite{mutegeki2020cnn} combined the CNN with LSTM, and the recognition accuracy on the UCI dataset reached 92\%.

Unfortunately, HAR research is completely dependent on the data quality of human activity datasets. Due to the inconsistencies in the difficulty of collecting different human activity data, such as walking data is easier to collect than the data of going up and downstairs, the data amount of some activities in the human activity dataset is always small, resulting in the human activity dataset extremely imbalanced. In this paper, we adopt GAN framework to generate sensor data and balance the human activity dataset with the generated activity data.

\subsection{Generative Adversarial Networks}
Generative Adversarial Networks (GAN) developed fast in recent years. Inspired by the original GAN, researchers have proposed many variants of GAN. WGAN \cite{arjovsky2017wasserstein} and its improved version WGAN-GP\cite{gulrajani2017improved} are proposed to solve the problem of mode collapse and vanishing gradients of the original GAN during training. Mirza M et al.\cite{mirza2014conditional} proposed CGAN to bring label information into the training process of GAN, so that GAN can generate data for specific classes. Radford et al.\cite{radford2015unsupervised} proposed DCGAN, they use deep convolutional neural networks to replace the MLP in the original GAN and replace the fully connected layer with a global pooling layer to reduce the amount of computation. Based on CGAN and DCGAN, Odena A et al.\cite{odena2017conditional} proposed ACGAN. When the discriminator of ACGAN discriminates the real data, it also classifies the real data. The results of the classification are feedback to the generator to improve the data quality. The researchers of BAGAN\cite{mariani2018bagan} point out that when ACGAN works on unbalanced and small datasets, it cannot effectively generate data for the minority classes. BAGAN improves the loss function of ACGAN to solve the self-contradiction between the two loss functions of ACGAN.
All of the above GANs are designed to generate high-quality generated data and stabilize the GAN training process. Meanwhile, GAN has gradually expanded from the field of image generation to many other applications, such as natural language processing\cite{yu2017seqgan}, speech synthesis\cite{kong2020hifi}, super-resolution reconstruction\cite{you2019ct}, text generation\cite{zhang2017adversarial}, and image inpainting\cite{demir2018patch}.

M.Alzantot et al.\cite{alzantot2017sensegen} proposed SenseGEN to generate activity data, which is the first application of GAN framework on human activity datasets. However, the discriminator and generator model of SenseGEN are trained separately, the generator model cannot learn from the feedback of the discriminator model, resulting in poor quality of the generated activity data. Wang J et al.\cite{wang2018sensorygans} proposed SensoryGANs to oversample human activity dataset and improve the performance of HAR models, but the generator model needs to switch different network structures for different human activities, in order to adapt to the data distribution features of various human activities, which makes SensoryGANs need to train multiple generators on the same activity dataset. Alharbi et al.\cite{alharbi2020synthetic} employed WGAN to generate activity sensor data. Considering the activity data is time-sequence data, they adopt RNN based model to build the generator and discriminator model, which makes their GAN structure difficult to train. Razvan Pascanu et al.\cite{pascanu2013difficulty} showed that the neural network architecture based on the RNN model has unstable convergence during training.
The generator and discriminator model of the proposed BSDGAN in this paper is built by 1D-transposed CNN and 1D-CNN. In the following sections, we will demonstrate that GANs built by pure convolutions can also generate data for different human activities.

\section{Balancing Sensor Data GAN}
\begin{figure}[htbp]
\centerline{\includegraphics{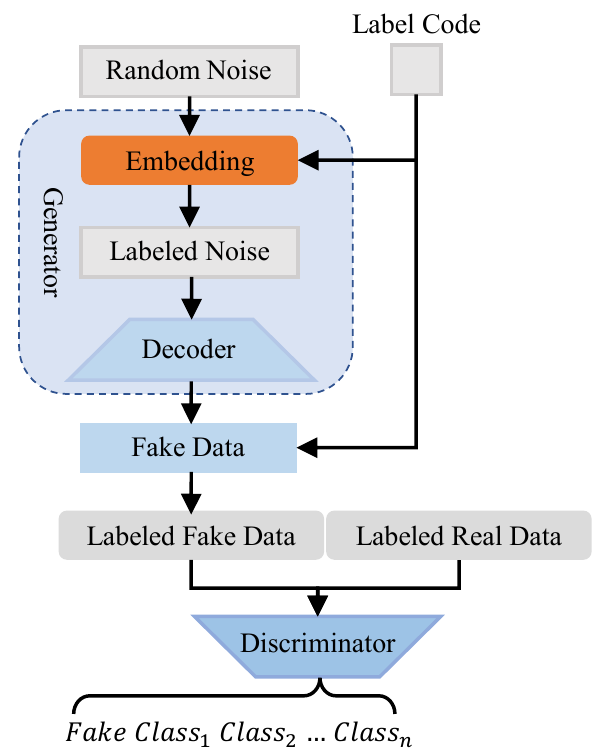}}
\caption{Architecture of proposed BSDGAN.}
\label{fig1}
\end{figure}

We dub our generative adversarial networks framework as Balancing Sensor Data GAN (BSDGAN). BSDGAN is a complete generative adversarial networks framework, consists of a generator model and a discriminator model.
Human activity datasets are usually extremely imbalanced, traditional GANs tend to generate sensor data for activity class with a large number of data samples, rather than the minority class data we need. To solve this problem, we adopt an autoencoder to initialize the training of BSDGAN, so that BSDGAN can learn the data features of all activity classes.
The architecture of BSDGAN is shown in Fig.~\ref{fig1}.

Randomly noise and label information are simultaneously fed into the generator model to generate fake sensor data. Subsequently, the fake sensor data with label information and the real sensor data are both fed into the discriminator, the discrimination results will be employed to improve the parameters of the generator model.

\subsection{Autoencoder Initialization}\label{AA}

Autoencoders can converge towards good solutions easily\cite{doersch2016tutorial}. We apply an autoencoder to initialize the GAN, let it close to a good solution at the initial stage, away from mode collapse. Furthermore, the encoder part of the autoencoder is adopted to infer the distribution of latent vectors for different human activity classes.

\begin{figure}[htbp]
\centerline{\includegraphics{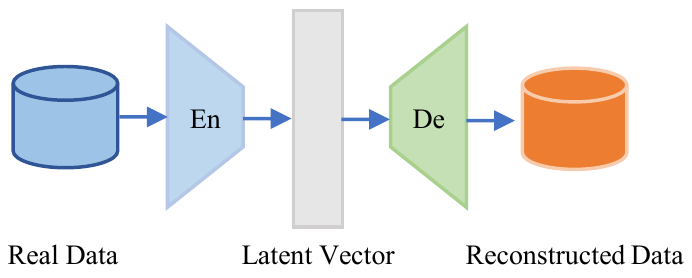}}
\caption{Autoencoder Initialization.}
\label{fig2}
\end{figure}

The initialization process of the autoencoder is shown in Fig.~\ref{fig2}. All sensor data in the human activity dataset will be used to train the autoencoder. Training process is performed by mean absolute error loss function minimization.

\subsection{Design of Generator Model}

\begin{figure}[htbp]
\centerline{\includegraphics[width=0.48\textwidth]{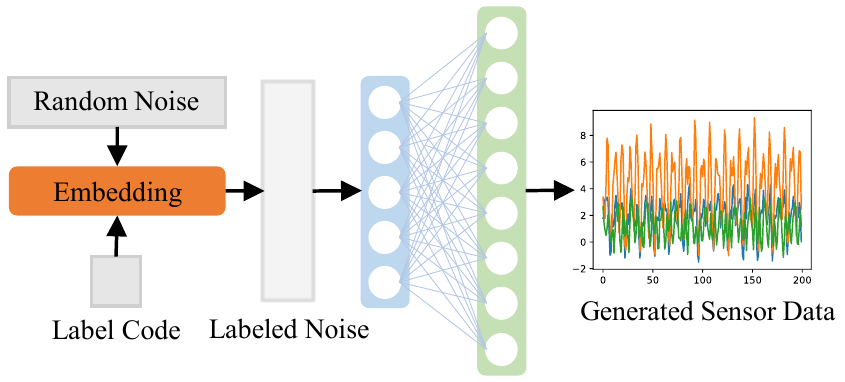}}
\caption{Design of generator model.}
\label{fig3}
\end{figure}

The goal of the generator model is to learn the data feature distribution of real sensor data $p(x|c_i)$ in specific activity class $c_i$, then generate high quality synthetic data $G(z|c_i)$.
The generator model of BSDGAN consists of an embedding layer and a pre-trained decoder, as shown in Fig.~\ref{fig3}. Random noise $z$ and a label information $c_i$ are fed into the embedding layer to generate labeled noise.
The pre-trained decoder shares the weight parameters and network architecture with the generator model. The labeled noise can be converted to the sensor data $G(z|c_i)$ in specific activity class by 1D-transposed convolutional neural network layer in the decoder. The weight parameters of decoder in generator can be updated during the process of adversarial training.

\subsection{Architecture of the Discriminator Model}
\begin{figure}[htbp]
\centerline{\includegraphics{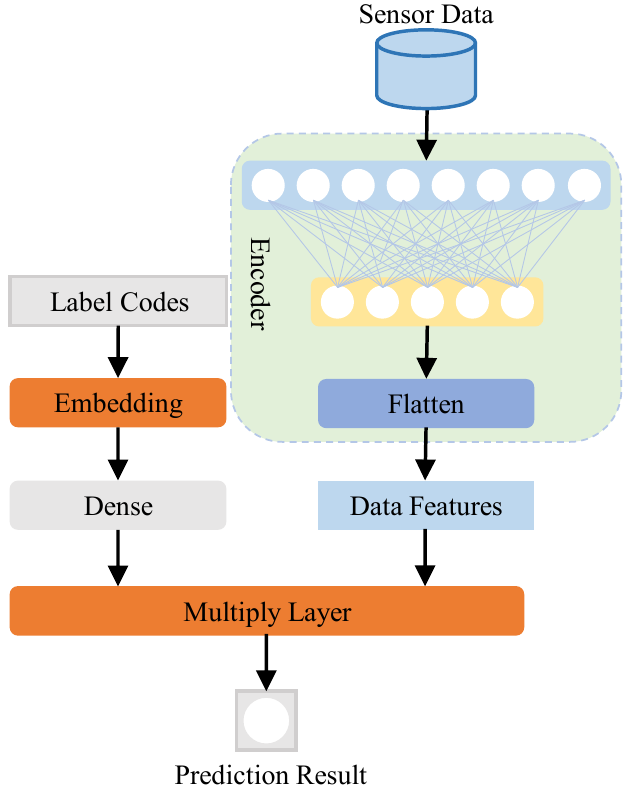}}
\caption{Architecture of discriminator model.}
\label{fig4}
\end{figure}

The discriminator model in BSDGAN is an extension of the pre-trained encoder.
The discriminator does not employ the entire architecture of the encoder, the last layer of the encoder is removed, and the output of the second-last layer (feature map) with the label information are fed into the embedding layer, generate a new dense vector.
The last layer of the discriminator model is a dense layer with $softmax$ as activation function. This layer is applied to output the discrimination result of the input data. If the discrimination result is the generated data, it outputs the label $fake$, and if the discrimination result is the real data, it outputs the corresponding activity label code $c_i$. The architecture of the discriminator model is shown in the Fig.~\ref{fig4}.

\subsection{Improved Loss Function}
The training process of GAN is the process of the discriminator against the generator. The goal of generator $G$ is to generate data that fools the discriminator, and the goal of discriminator $D$ is to distinguish real data from generated data. Therefore, the loss function of original GAN is defined as:
\begin{multline}
    \mathop{min}_{G}\mathop{max}_{D}V(D, G)=\\\mathbb{E}_{{x_r}\sim{X_r}}[logD({x_r})]+
    \mathbb{E}_{{x_g}\sim{X_g}}[log(1-D({x_g}))]
\end{multline}
where $x_r$ is the real sensor data, $X_r$ is the real data distribution, $x_g=G(z)$ is the generated sensor data, and $X_g$ is the generated data distribution.

The original GAN cannot generate data for specified classes. Conditional GAN (CGAN) adds constraints on the basis of the original GAN, enabling it to generate synthetic data of specified classes. The loss function of CGAN is defined as follows:
\begin{multline}
    \mathop{min}_{G}\mathop{max}_{D}V(D, G)=\mathbb{E}_{{x_r}\sim{X_r}}[logD({x_r|y})]+\\
    \mathbb{E}_{{x_g}\sim{X_g}}[log(1-D({x_g|y}))]
\end{multline}
where $y$ is the conditional constraint. It is combined with input noise $z$ to form a joint hidden layer representation. The discriminator discriminates between $x_g$ and $x_r$ based on the condition $y$.

To make the training process of GAN more stable, WGAN improves the loss function of GAN. The loss function of GAN is based on Jensen-Shannon divergence, which makes the training process of GAN difficult. WGAN uses Wasserstein Distance instead of JS divergence to measure the difference between generated data and real data. Wasserstein Distance is defined as:
\begin{equation}
    W(X_r, X_g) = \mathop{inf}_{\gamma \in \prod(X_r, X_g)}\mathbb{E}_{(x_r, x_g)}\sim{\gamma}{[\left\|x_r-x_g\right\|]}
\end{equation}
where $\prod({X_r, X_g)}$ represents the all joint distributions between $X_r$ and $X_g$. However, wasserstein distance is difficult to deal with in practice, researchers employed Kantorovich-Rubinstein duality as an alternative:
\begin{equation}
    W(X_r, X_g) = {\mathop{sup}_{{\left\|D\right\|}_L \leq 1}}({\mathbb{E}_{x_r \sim{X_r}}[D(x_r)]}-\mathbb{E}_{x_g\sim{X_g}}[D(x_g)])
\end{equation}
where ${\left\|D\right\|}_L \leq 1$ means discriminator $D$ follows the 1-Lipschitz function. So the objective of WGAN is obtained as:
\begin{equation}
     W(X_r, X_g)= \mathbb{E}_{x_r\sim{X_r}}[D(x_r)]-\mathbb{E}_{x_g\sim{X_g}}[D(x_g)]
\end{equation}

WGAN directly adopts weight clipping when dealing with 1-Lipschitz constraints, which results in the parameters of the discriminator focusing on the maximum and minimum values, easily leading to gradient vanishing and gradient explosion. WGAN-GP employed gradient penalty to improve the loss function of the discriminator model. The loss function of WGAN-GP is defined as:
\begin{equation}
    \begin{aligned}
        \mathop{min}_{G}&\mathop{max}_{D}V(D, G)=\mathbb{E}_{{x_g}\sim{X_g}}[D(x_g)]-
        \mathbb{E}_{{x_r}\sim{X_r}}[D({x_r})]\\
        &-\lambda\mathbb{E}_{{\hat{x}\sim{\hat{X}}}}[(\Vert {\triangledown_{\hat{x}}}D(\hat{x}) \Vert_2-1)^2]
    \end{aligned}
\end{equation}
where $\hat{x}=\alpha {x_r} + (1-\alpha){x_g}$, $\alpha$ is in the range $0$ to $1$. $\lambda$ is a hyperparameter of the penalty extent.

In this paper, we combine the improvements of WGAN-GP on the original GAN with the conditional constraints of CGAN. The loss function of BSDGAN is defined as follows:
\begin{equation}
    \begin{aligned}
        \mathop{min}_{G}&\mathop{max}_{D}V(D, G)=\\
        &\mathbb{E}_{{x_r}\sim{X_r},{y_r}\sim{Y_r}}[logD({x_r}|{y_r})]+\\
        &\mathbb{E}_{{x_g}\sim{X_g}, {y_g}\sim{Y_g}}[log(1-D(G({x_g| y_g})|{y_g}))]+\\
        &\mathbb{E}_{{x_r}\sim{X_r}, {y_{wrong}}\sim{Y_{wrong}}}[log(1-D({x_r}|y_{wrong}))]-\\
        &\lambda\mathbb{E}_{{\hat{x}\sim{\hat{X}}}, {y_r}\sim{Y_r}}[(\Vert {\triangledown_{\hat{x}|y_r}}D(\hat{x}|y_r) \Vert_2-1)^2]
    \end{aligned}
\end{equation}
where $y_r$ is the real label and $Y_r$ is the set of all real data labels. In an imbalanced dataset, ground truth labels randomly sampled from the dataset are still imbalanced. Therefore, we refer to BAGAN and randomly select a label $y_g$ for each generated data from the balanced label set $Y_g$. To enhance the learning of class information from the real dataset, we add an additional cross-entropy loss for misclassified cases, and the misclassified data label is set to $y_{wrong}$.
\subsection{Pseudocode of Algorithm}
The pseudocode of the dataset balance using BSDGAN is shown in Algorithm \ref{alg:Framwork}. During the balancing process, BSDGAN takes the data amount of the largest class as the balancing criterion and oversamples each human activity class. When the amount of data for all activity classes is greater than the balance standard, the balancing process ends, this means that the dataset reaches a balanced state.

\begin{algorithm}[htb] 
\caption{The process of human activity dataset balance} 
\label{alg:Framwork} 
\begin{algorithmic}[1] 
\REQUIRE ~~ 
random noise $z$; real dataset $x_{real}$;
\ENSURE ~~ 
the balanced dataset $x_{b}$
\STATE Split the real dataset $x_{real}$ into real sensor data $x_r$ and real data label $y_r$;
\STATE Import the trained generator model $G$ and discriminator model $D$;
\STATE Calculate the amount of data for each class, set the maximum value to $N_c^{max}$;
\FOR{$i$=1 to $len(classes)$}
    \STATE Set the data volume of $i^{th}$ class to $N_c^i$;
    \WHILE{$N_c^i < N_c^{max}$}
        \STATE Generate the fake data $x_g^i$ for the specified class $c_i$ with $G(z, c_i)$;
        \STATE Employ discriminator model $D$ to verify the output of generator as $c_g$;
    \IF{$c_i = c_g$}
        \STATE Add the generated data $x_g^i$ to the fake dataset $x_g$;
        \STATE $N_c^i$++;
    \ENDIF
    \ENDWHILE
\ENDFOR
\STATE Combine the generated dataset $x_g$ with the real dataset $x_r$ to get the balanced dataset $x_b$;
\RETURN $x_{b}$
\end{algorithmic}
\end{algorithm}

\section{Experimentation and Evaluation}

\subsection{Datasets}
\begin{figure}[htbp]
\centerline{\includegraphics[width=0.5\textwidth]{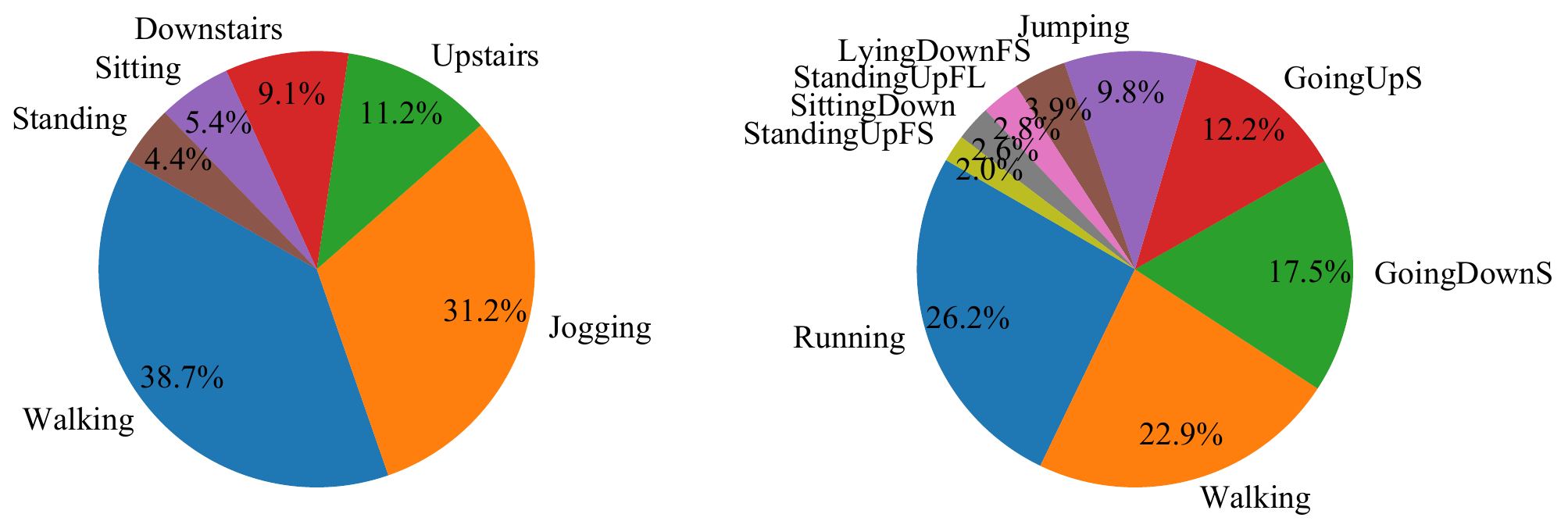}}
\caption{Sensor Data Distribution of WISDM (Left) and Unimib-SHAR (Right).}
\label{fig_pie}
\end{figure}

In this paper, we use two public human activity datasets, WISDM\cite{kwapisz2011activity} and Unimib-SHAR\cite{micucci2017unimib} to train and evaluate our BSDGAN. 
WISDM contains 6 types of human activity classes with a total of 1,098,207 rows of data (54,901 human activity instances) collected by 36 users. Each row of data contains three-axis acceleration values and timestamp information. The Unimib-SHAR (ADL) dataset contains 9 daily living activities with a total of 7579 instances of human activity data, each activity instance contains 151 rows of raw data. Each row of data contains three-axis acceleration data, timestamp, and raw signal amplitude, collected by 30 volunteers aged between 18 and 60.
These two human activity datasets both suffer from severe data imbalance, the data distributions of the two datasets are shown in Fig.~\ref{fig_pie}.

\subsection{Training Process}
In this paper, the framework of all experiments is Tensorflow 2.5.0 based on Python. The proposed BSDGAN is evaluated on a laptop computer with an NVIDIA RTX2060. We use Adam algorithm as the optimizer for generator model and discriminator model. The learning rate is 0.0002, beta1 and beta2 are set to 0.5 and 0.9. The size of batch is 128, default latent vector is 100 dimensions. We train 100 epochs on two human activity datasets, each epoch takes 88s on WISDM and 15s on Unimib-SHAR.

\begin{figure}[htbp]
    \centering
    \subfigure[]{
        \label{fig_loss(a)}
        \includegraphics[width=6.7cm]{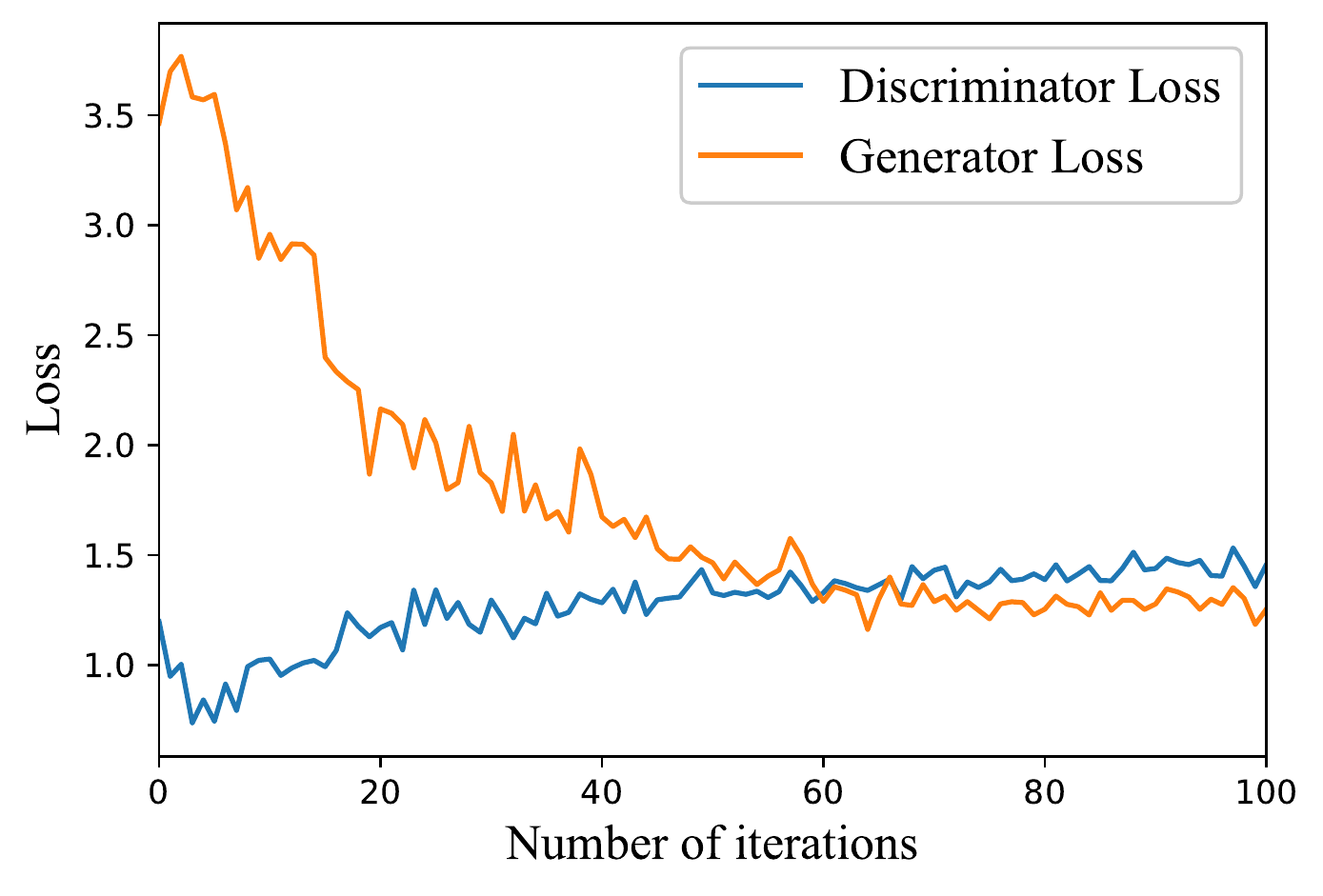}}
    \\
    \subfigure[]{
        \label{fig_loss(b)}
        \includegraphics[width=6.7cm]{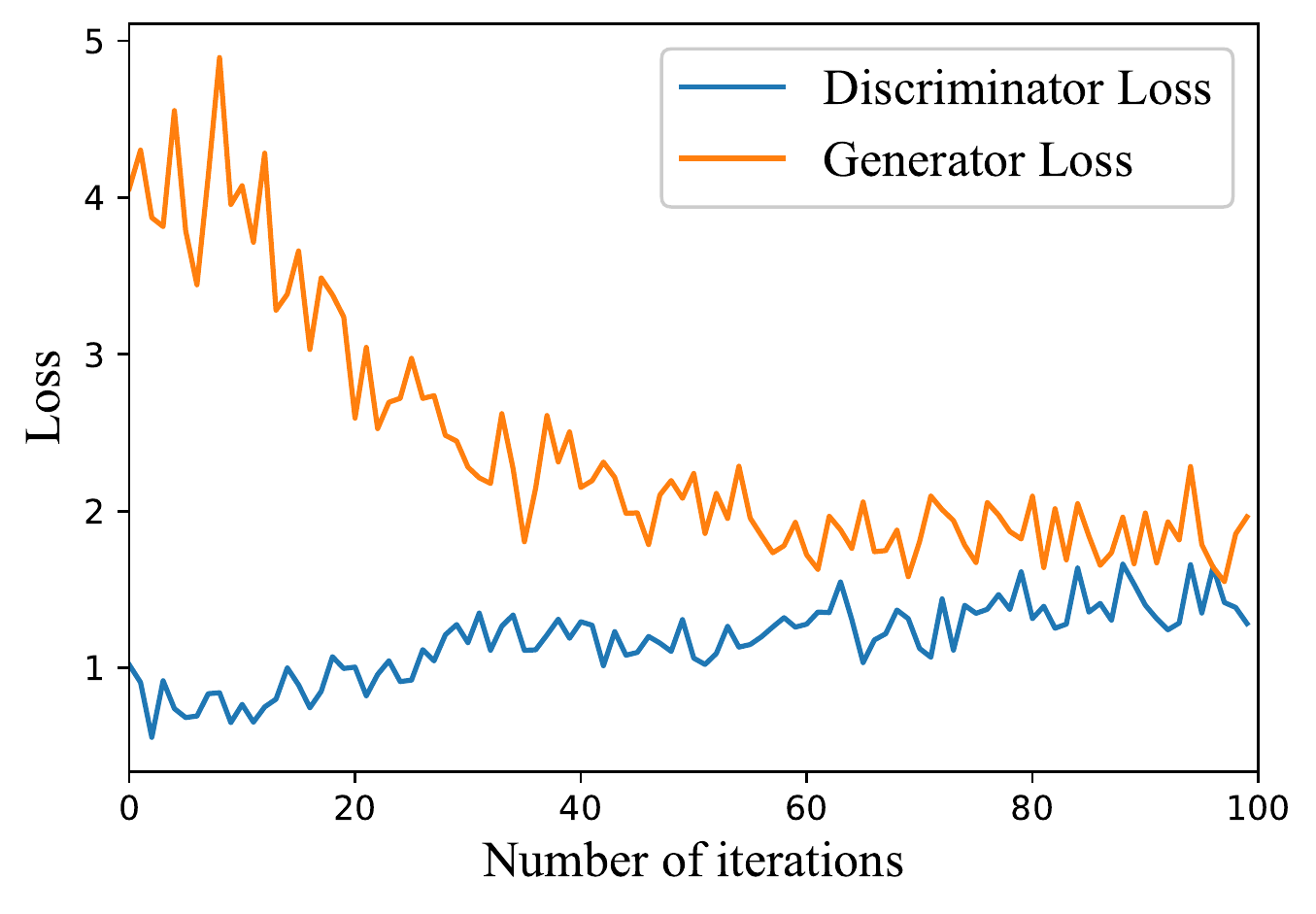}}
    \caption{The generator loss and discriminator loss during training. (a) Loss Comparison on WISDM; (b) Loss Comparison on Unimib-SHAR.}
    \label{fig_loss}
\end{figure}

During the training process of BSDGAN on the WISDM dataset and UNIMIB dataset, the loss values of generator and discriminator are shown in Fig.~\ref{fig_loss}. Benefit from the pre-trained encoder and decoder, BSDGAN can quickly achieve the Nash equilibrium of the generator and discriminator on both datasets. This result demonstrates that BSDGAN can learn the data distribution of real sensor data on imbalanced datasets and generate high-quality synthetic data.

The data distribution comparison of WISDM and Unimib before and after data balance is shown in Table ~\ref{tab_wis} and Table ~\ref{tab_uni}. We adopt Algorithm \ref{alg:Framwork} to automatically balance the two human activity datasets, and the amount of data for all activities reach a substantially consistent state after the data balance. In the following section, we compare the performance of various human activity classifiers on the above activity datasets before and after balancing.

\begin{table*}
    \centering
    \caption{Data amount of WISDM dataset.}
    \label{tab_wis}
    \setlength{\tabcolsep}{3mm}{
        \begin{tabular}{ccccccc}
        \hline
        State & Walking & Jogging & Upstairs & Downstairs & Sitting & Standing \\
        \hline
        imbalance & 21220 & 17104 & 6146 & 5020 & 2992 & 2419 \\
        balance & 21220 & 21220 & 21224 & 21220 & 21220 & 21223 \\
        \hline
        \end{tabular}
    }
\end{table*}

\begin{table*}
    \centering
    \caption{Data amount of Unimib-SHAR dataset.}
    \label{tab_uni}
        \begin{tabular}{cccccccccc}
        \hline
        State & Running & Walking & GoingDownS & GoingUpS & Jumping & LyingDownFS & StandingUpFL & SittingDown & StandingUpFS \\
        \hline
        imbalance & 1985 & 1738 & 1324 & 921 & 746 & 296 & 216 & 200 & 153 \\
        balance & 1985 & 1989 & 1990 & 1985 & 1992 & 1988 & 1990 & 1988 & 1991 \\
        \hline
        \end{tabular}
\end{table*}

\subsection{Quality Assessment of Generated Data}
This paper employed the Frechet Inception Distance\cite{heusel2017gans} Score (FID) to evaluate the quality of the generated activity data. The FID score is usually employed to calculate the distance between the data feature distribution of the generated data and the data feature distribution of the real data. Frechet Inception Distance is defined as:
\begin{equation}
	\begin{aligned}
		FID({x_r},{x_g}) = & ||{\mu_r} - {\mu_g}|{|^2} + 
		\\&{T_r}({\sum}_r + {\sum}_g - 2{({{\sum}_r}{{\sum}_g})^{\frac{1}{2}}})
	\end{aligned}
\end{equation}
where $\mu_r$ is the mean of the real data features, $\mu_g$ is the mean of the generated data features, ${\sum}_r$ is the covariance matrix of the real data features, ${\sum}_g$ is the covariance matrix of the generated data features.

All FID scores are calculated from the real data of the validation set and the data generated by the generator model. The lower the FID score of the generated data, the higher the data quality. In this paper, we define the FID score between the real data of the training set and the data of the validation set as the best score, the FID score between the data reconstructed by the autoencoder and the data of the validation set as the worst result. We deployed CGAN, ACGAN, BAGAN, and the proposed BSDGAN on the above activity datasets, evaluating the data quality of generated data by these GAN frameworks. On WISDM, the comparison results of the generated data quality for each activity class are shown in Table \ref{tab_wis_fid_compare}, and the comparison results of the generated data quality on Unimib-SHAR-ADL are shown in Table \ref{tab_unimib_fid_compare}.
\begin{table*}
	\renewcommand\arraystretch{1.2}
    \centering
    \caption{Comparison of FID for generated data by different GANs on WISDM.}
    \label{tab_wis_fid_compare}
    \setlength{\tabcolsep}{5mm}{
        \begin{tabular}{ccccccc}
        \hline
        Activity & Autoencoder & CGAN & ACGAN & BAGAN & BSDGAN & Real Data \\
        \hline
        Jogging & 298.11 & 240.21 & 197.66 & 165.35 & \pmb{119.62} & 87.23\\
        Walking & 129.28 & 103.89 & 85.45 & 73.28 & \pmb{68.14} & 60.15\\
        UpStairs & 150.82 & 138.77 & 90.98 & 96.19 & \pmb{72.89} & 42.31\\
        DownStairs & 97.37 & 93.75 & 86.69 & 78.01 & \pmb{60.72} & 51.28\\
        Sitting & 247.50 & 214.87 & 152.97 & 103.29 & \pmb{96.18} & 66.34\\
        Standing & 151.06 & 134.94 & 88.56 & 83.64 & \pmb{54.58} & 28.84\\
        \hline
        \end{tabular}
    }
\end{table*}

\begin{table*}
	\renewcommand\arraystretch{1.2}
    \centering
    \caption{Comparison of FID for generated data by different GANs on Unimib-SHAR.}
    \label{tab_unimib_fid_compare}
    \setlength{\tabcolsep}{5mm}{
        \begin{tabular}{ccccccc}
        \hline
        Activity & Autoencoder & CGAN & ACGAN & BAGAN & BSDGAN & Real Data \\
        \hline
        StandingUpFs & 148.65 & 105.68 & 75.65 & 63.81 & \pmb{48.27} & 28.81\\
        StandingUpFs & 87.92 & 79.93 & 56.54 & 53.31 & \pmb{43.90} & 31.02\\
        Walking & 154.09 & 127.37 & 82.03 & 78.34 & \pmb{69.73} & 52.81\\
        Running & 147.92 & 132.59 & 103.45 & 83.32 & \pmb{78.16} & 52.75\\
        GoingUpS & 131.51 & 106.39 & 83.21 & 74.98 & \pmb{64.82} & 48.32\\
        Jumping & 91.43 & 84.87 & 73.18 & 68.40 & \pmb{63.85} & 52.53\\
        GoingDownS & 113.51 & 96.23 & 89.05 & 85.21 & \pmb{61.14} & 57.31\\
        LingDownFS & 130.16 & 76.76 & 62.34 & 56.36 & \pmb{46.28} & 31.76\\
        SittingDown & 92.61 & 85.64 & 76.35 & 68.52 & \pmb{41.73} & 38.74\\
        \hline
        \end{tabular}
    }
\end{table*}

\begin{table*}
	\renewcommand\arraystretch{1.2}
    \centering
    \caption{Classification accuracy on WISDM before balance.}
    \label{tab_wis_before_balance}
    \setlength{\tabcolsep}{5mm}{
        \begin{tabular}{ccccccc}
        \hline
        Activity & KNN & RF & DT & CNN & LSTM & CNN-LSTM \\
        \hline
        Jogging & 0.9208 & 0.9756 & 0.8378 & 0.9936 & 0.9777 & 0.9988\\
        Walking & 0.9669 & 0.9906 & 0.7819 & 0.9826 & 0.9885 & 0.9932\\
        UpStairs & 0.1899 & 0.3922 & 0.3270 & 0.7589 & 0.9345 & 0.9777\\
        DownStairs & 0.0960 & 0.1269 & 0.3182 & 0.7566 & 0.9216 &0.9489\\
        Sitting & 0.9861 & 0.9653 & 0.9722 & 0.9844 & 0.9723 & 0.9896\\
        Standing & 0.9885 & 0.9358 & 0.9128 & 0.9702 & 0.9771 & 0.9702\\
        Accuracy & 0.7849 & 0.8333 & 0.7199 & 0.9393 & 0.9714 & 0.9878\\
        \hline
        \end{tabular}
    }
\end{table*}

\begin{table*}
	\renewcommand\arraystretch{1.2}
    \centering
    \caption{Classification accuracy on WISDM after balance.}
    \label{tab_wis_after_balance}
    \setlength{\tabcolsep}{5mm}{
        \begin{tabular}{ccccccc}
        \hline
        Activity & KNN & RF & DT & CNN & LSTM & CNN-LSTM \\
        \hline
        Jogging & 0.9310 & 0.9714 & 0.9924 & 0.9927 & 0.9959 & 0.9994\\
        Walking & 0.9687 & 0.9899 & 0.9866 & 0.9817 & 0.9951 & 0.9969\\
        UpStairs & 0.7739 & 0.7835 & 0.7828 & 0.9315 & 0.9678 & 0.9637\\
        DownStairs & 0.7736 & 0.7527 & 0.7897 & 0.9138 & 0.9669 & 0.9782\\
        Sitting & 0.9979 & 0.9931 & 0.9953 & 0.9653 & 0.9826 & 0.9340\\
        Standing & 0.9783 & 0.9909 & 0.9976 & 0.9748 & 0.9885 & 1.0\\
        Accuracy & 0.9039 & 0.9138 & 0.9241 & 0.9600 & 0.9887 & 0.9891\\
        \hline
        \end{tabular}
    }
\end{table*}

\begin{table*}
	\renewcommand\arraystretch{1.2}
    \centering
    \caption{Classification accuracy on Unimib-SHAR before balance.}
    \label{tab_uni_before_balance}
    \setlength{\tabcolsep}{5mm}{
        \begin{tabular}{ccccccc}
        \hline
        Activity & KNN & RF & DT & CNN & LSTM & CNN-LSTM \\
        \hline
        StandingUpFs & 0.9091 & 1.0 & 0.8788 & 0.9394 & 0.6970 & 0.9697\\
        StandingUpFL & 0.5600 & 0.3800 & 0.4800 & 0.6200 & 0.5600 & 0.7800\\
        Walking & 0.9809 & 0.7350 & 0.7814 & 0.9481 & 0.9891 & 0.9836\\
        Running & 0.9262 & 0.8626 & 0.8295 & 0.9873 & 0.9924 & 1.0\\
        GoingUpS & 0.7486 & 0.2131 & 0.5573 & 0.6503 & 0.8798 & 0.9016\\
        Jumping & 0.7600 & 0.5667 & 0.7333 & 0.9533 & 0.9600 & 0.9933\\
        GoingDownS & 0.7303	& 0.3942 & 0.6349 & 0.8174 & 0.9544 & 0.9627\\
        LyingDownFS & 0.7018 & 0.1930 & 0.5439 & 0.7193 & 0.5965 & 0.8245\\
        SittingDown & 0.8140 & 0.3488 & 0.6512 & 0.6511 & 0.8605 & 0.7906\\
        Accuracy & 0.8463 & 0.5970 & 0.7183 & 0.8740 & 0.9294 & 0.9571\\
        \hline
        \end{tabular}
    }
\end{table*}

\begin{figure}[htbp]
\centerline{\includegraphics[width=0.5\textwidth]{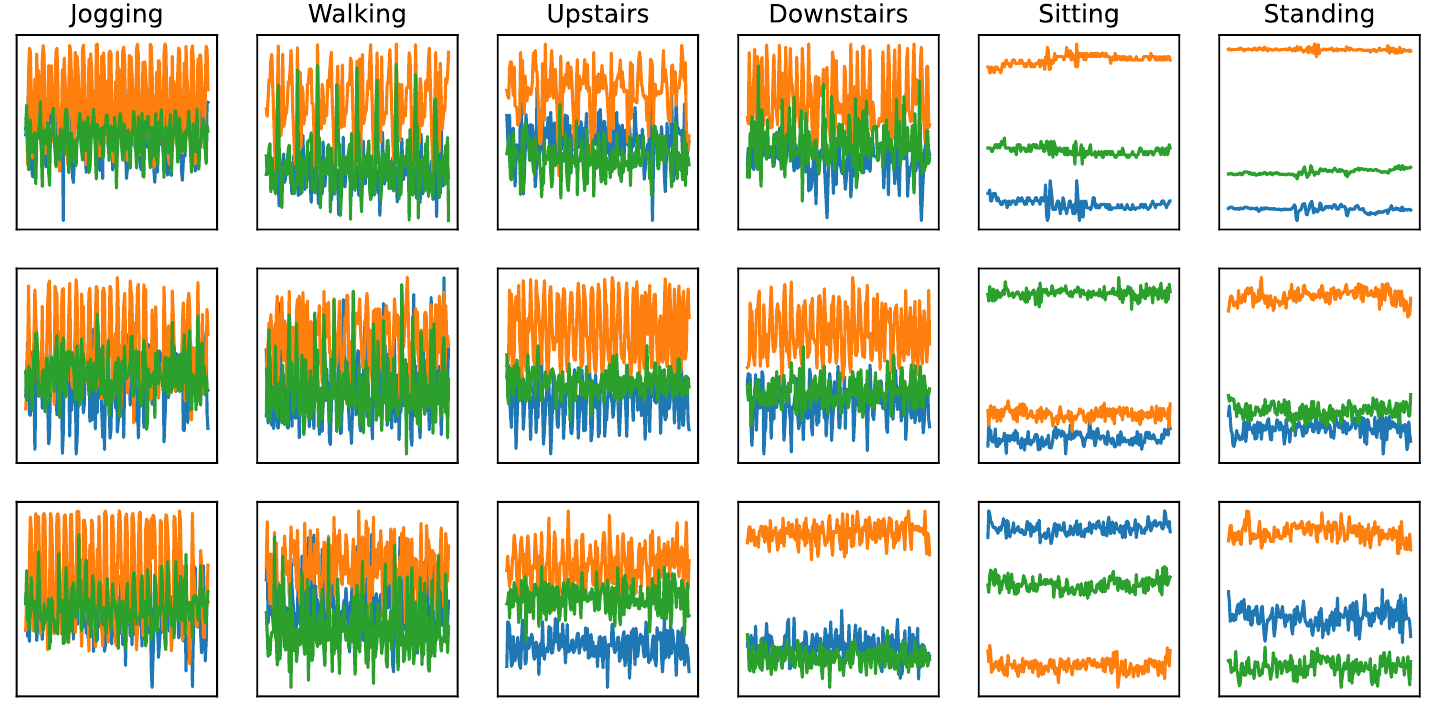}}
\caption{The generated sensor data on WISDM.}
\label{fig5}
\end{figure}

\begin{figure}[htbp]
\centerline{\includegraphics[width=0.5\textwidth]{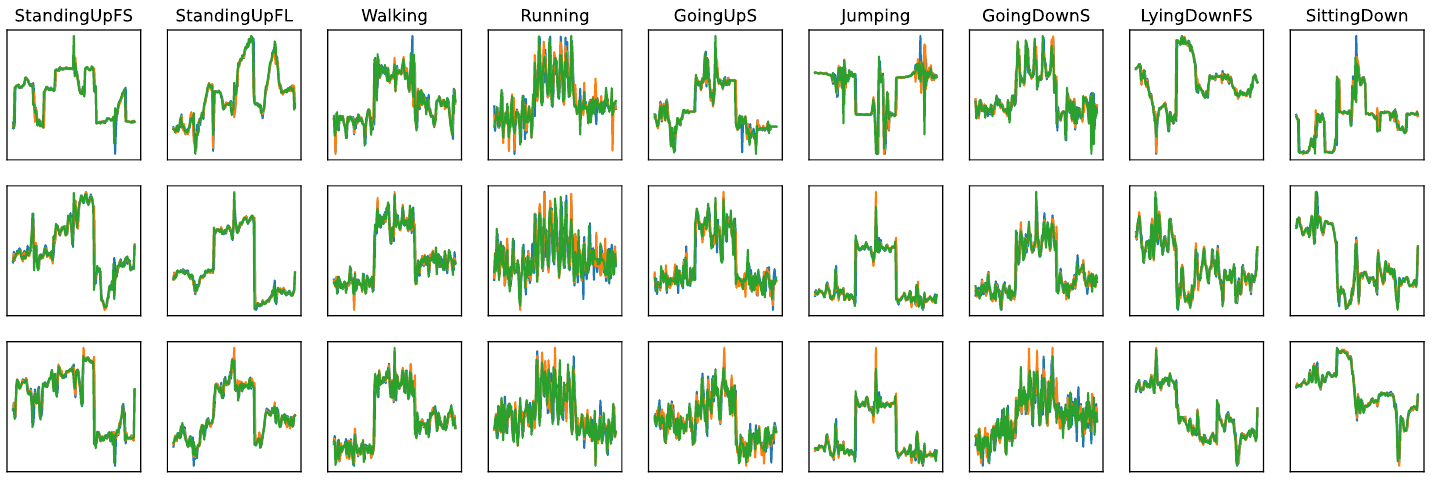}}
\caption{The generated sensor data on Unimib-SHAR.}
\label{fig6}
\end{figure}

\begin{table*}
	\renewcommand\arraystretch{1.2}
    \centering
    \caption{Classification accuracy on Unimib-SHAR after balance.}
    \label{tab_uni_after_balance}
    \setlength{\tabcolsep}{5mm}{
        \begin{tabular}{ccccccc}
        \hline
        Activity & KNN & RF & DT & CNN & LSTM & CNN-LSTM \\
        \hline
        StandingUpFs & 0.9831 & 0.9177 & 0.9564 & 0.9758 & 0.9758 & 0.9903\\
        StandingUpFL & 0.9689 & 0.9019 & 0.9426 & 0.9833 & 0.9498 & 0.9904\\
        Walking & 0.9923 & 0.9872 & 0.8619 & 0.9284 & 0.9821 & 0.9974\\
        Running & 0.9340 & 0.9902 & 0.8973 & 0.9829 & 0.9756 & 0.9976\\
        GoingUpS & 0.8862 & 0.5932 & 0.7530 & 0.8838 & 0.9540 & 0.9515\\
        Jumping & 0.9147 & 0.7360 & 0.8640 & 0.9760 & 0.9840 & 0.9893\\
        GoingDownS & 0.8392 & 0.7139 & 0.7163 & 0.9125 & 0.9622 & 0.9835\\
        LyingDownFS & 0.9457 & 0.8829 & 0.9229 & 0.9486 & 0.9514 & 0.9829\\
        SittingDown & 0.9794 & 0.9383 & 0.9537 & 0.9820 & 0.9589 & 0.9769\\
        Accuracy & 0.9374 & 0.8500 & 0.8727 & 0.9522 & 0.9659 & 0.9844\\
        \hline
        \end{tabular}
    }
\end{table*}
This paper visualizes the generated activity data on two activity datasets, as shown in Fig.~\ref{fig5} and Fig.~\ref{fig6}. The data visualization of different human activities is obviously different, the visual view of sitting and standing is relatively flat, while the view of other human activities, such as Jogging and Walking, fluctuates greatly. Besides, there are differences between the generated data for each human activity, which conforms to the data features of real human activities.

\subsection{Usability of Synthetic Data}
The ultimate goal of BSDGAN is to augment human activity datasets with generated data and improve the classification accuracy of human activity recognition models.
The generator model of BSDGAN is adopted to generate 4116 Jogging instances, 15078 UpStairs instances, 16220 DownStairs instances, 18230 Sitting instances, and 18804 Standing instances on the WISDM dataset. On the Unimib-SHAR dataset, 251 Walking instances, 666 GoingDownS instances, 1064 GoingUpS instances, 1246 Jumping instances, 1692 LyingDownFS instances, 1774 StandingUpFL instances, 1788 SittingDownS instances, and 1838 StandingUpFS instances are generated. The generation of these data is performed automatically by Algorithm \ref{alg:Framwork} to ensure that the activity dataset is balanced.
We adopted K-Nearest Neighbors (KNN), Random Forest (RF), Decision Tree (DT), three traditional machine learning methods and Convolutional Neural Network (CNN), Long Short-Term Memory (LSTM), CNN-LSTM three classical deep learning methods as classifiers for human activity recognition to evaluate the human activity dataset before and after the data balance.

The parameters of the above three human activity recognition classifiers based on machine learning methods are selected by grid search. The classifier based on CNN consists of a 1D-Convolutional Layer with 64 filters, kernel size is set to 1$\times$3, a 1D-MaxPooling layer and a dense layer with $softmax$ function as output layer. The LSTM based classifier contains a LSTM layer with 100 units, and the CNN-LSTM based classifier is the combination of the previous two models. The convolutional layers transmit the calculated data features to LSTM layers, achieve the best classification accuracy.

The performance of the above six human activity recognition classifiers before and after the balance on the WISDM dataset is shown in Table~\ref{tab_wis_before_balance} and Table~\ref{tab_wis_after_balance}, and the performance on the Unimib-SHAR dataset before and after the balance is shown in Table~\ref{tab_uni_before_balance} and Table~\ref{tab_uni_after_balance}. The classification accuracy of traditional machine learning methods has been greatly improved after the dataset is balanced. For example, the accuracy of Decision Tree on WISDM has increased from 71.99\% to 92.41\%, and the accuracy of Random Forest on Unimib-SHAR has increased from 59.70\% to 85.00\%.

The deep learning methods also solved the problem of poor recognition accuracy for minority activity classes. For example, the accuracy of CNN for upstairs in WISDM has increased from 75.89\% to 93.15\%, and the accuracy of LSTM for LyingDownfFS on Unimib-SHAR has increased from 59.65\% to 95.14\%. The above experimental results show that our BSDGAN can effectively generate high-quality human activity data, and the human activity dataset balanced with the generated data can obviously improve the accuracy of human activity recognition classifiers.

\section{Conclusion and future work}
In this paper, we use generative adversarial network framework to generate human activity sensor data, and the generated sensor data are adopted to balance the human activity dataset. We employ an autoencoder to give the GAN framework prior knowledge for all activity classes and help stabilize the GAN training process. Besides, we add conditional constraints that enable the GAN framework to generate activity data for target human activity classes.
Our experiments on two public human activity datasets show that the performance of HAR classifiers all significantly improved after the dataset is balanced, and solves the problem that the recognition accuracy of deep learning method-based HAR classifiers is poor for the minority activity classes.

In the future, we will conduct experiments on more public human activity datasets and investigate the possibility of our BSDGAN in generating human activity data for designated persons. Besides, We will work on using the proposed BSDGAN to generate enough samples of human fall data and actually deploy it in healthcare.


\bibliographystyle{IEEEtran}
\bibliography{IEEEabrv,mylib}
\end{document}